\pdfoutput=1
\documentclass[11pt,dvipsnames]{article}

\usepackage{microtype}
\usepackage{graphicx}
\usepackage{subfigure}
\usepackage{booktabs} % for professional tables

\usepackage[final]{acl}

\usepackage{times}
\usepackage{latexsym}

\usepackage[T1]{fontenc}
\usepackage[utf8]{inputenc}

\usepackage{microtype}

\usepackage{inconsolata}

\usepackage{graphicx}

\usepackage{amsmath}
\usepackage{amssymb}
\usepackage{mathtools}
\usepackage{amsthm}
\usepackage{algorithm}
\usepackage{algorithmic}
\usepackage{multirow}
\usepackage{bbm}
\usepackage{xcolor}
\usepackage{tcolorbox}
\newcommand{\hquad}{\hspace{0.5em}} 
\newcommand{\fixedwidth}{{2.0cm}} 
\newcommand{\llmmarker}[1]{\tcbox[colback=SkyBlue!70, colframe=SkyBlue!70, 
    arc=5pt, boxrule=0pt, left=2pt, right=2pt, top=0pt, bottom=0pt, 
    on line, boxsep=1pt]{\makebox[\fixedwidth][c]{#1}}}
\newcommand{\humanmarker}[1]{\tcbox[colback=Salmon!70, colframe=Salmon!70, 
    arc=5pt, boxrule=0pt, left=2pt, right=2pt, top=0pt, bottom=0pt, 
    on line, boxsep=1pt]{\makebox[\fixedwidth][c]{#1}}}

\title{ExaGPT: Example-Based Machine-Generated Text Detection\\for Human Interpretability}

\author{Ryuto Koike$^1$ \hquad Masahiro Kaneko$^{2}$ \hquad Ayana Niwa$^{2}$ \hquad Preslav Nakov$^{2}$ \hquad Naoaki Okazaki$^{1,3,4}$ \\
  $^1$Institute of Science Tokyo \quad $^2$MBZUAI \quad $^3$AIST \quad $^4$NII LLMC\\
  \texttt{\{ryuto.koike@nlp., okazaki@\}comp.isct.ac.jp}\\
  \texttt{\{masahiro.kaneko, ayana.niwa, preslav.nakov\}@mbzuai.ac.ae}
  \\}

\begin{document}
\maketitle
\begin{abstract}
  Detecting texts generated by Large Language Models (LLMs) could cause grave mistakes due to incorrect decisions, such as undermining students' academic dignity.
  LLM text detection thus needs to ensure the interpretability of the decision, which can help users judge how reliably correct its prediction is. 
  When humans verify whether a text is human-written or LLM-generated, they intuitively investigate which of them it shares more similar spans with.
  However, existing interpretable detectors are not aligned with the human decision-making process and fail to offer evidence that users easily understand. 
  To bridge this gap, we introduce \textbf{ExaGPT}, an interpretable detection approach grounded in the human decision-making process for verifying the origin of a text.
  ExaGPT identifies a text by checking whether it shares more similar spans with human-written vs. with LLM-generated texts from a datastore.
  This approach can provide similar span examples that contribute to the decision for each span in the text as evidence.
  Our human evaluation demonstrates that providing similar span examples contributes more effectively to judging the correctness of the decision than existing interpretable methods.
  Moreover, extensive experiments in four domains and three generators show that ExaGPT massively outperforms prior interpretable detectors by up to +37.0 points of accuracy at a false positive rate of 1\%.
% We will release our code. 
% \footnote{\url{https://www.github.com/ryuryukke/ExaGPT}}
\end{abstract}

\section{Introduction}
LLMs can yield human-like texts in response to various textual instructions \citep{chatgpt,touvron2023llama}.
Ironically, the powerful generative capability has resulted in various misuses of LLMs, such as cheating on student homework assignments and mass-producing fake news \citep{tang2023science,wu2023survey}. Such abuse of LLMs has sparked the demand for discerning LLM-generated texts from human-written ones. 

% Recent studies have developed LLM-generated text detectors with promising performance \citep{su2023detectllm,Koike:OUTFOX:2024,hans2024spotting,verma2024ghostbuster,emi2024technicalreportpangramaigenerated}.

\begin{figure}[t]
 \begin{center}
 \small
  \centering\includegraphics[width=\columnwidth]{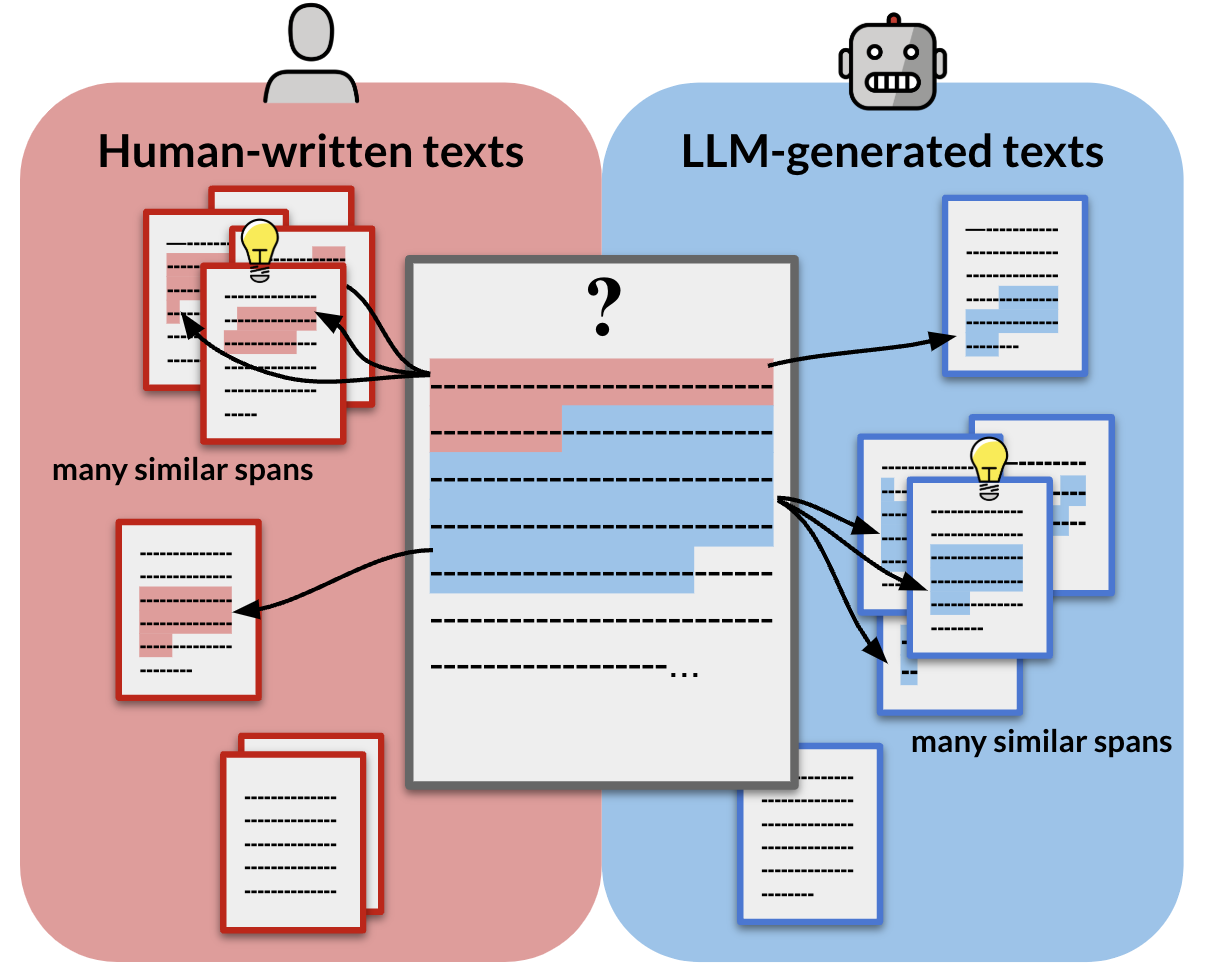}
  \caption{Identifying the author of a text (human vs. LLM) by examining if it shares more similar spans, including verbatim overlaps and semantically similar spans, with human-written vs. LLM-generated texts.}
  \label{human_intuitive_identification}
 \end{center}
\end{figure}

Recent studies have developed LLM-generated text detectors with promising performance \citep{su2023detectllm,Koike:OUTFOX:2024,hans2024spotting,verma2024ghostbuster,emi2024technicalreportpangramaigenerated}.

While LLM text detection can help prevent potential misuse of LLMs, misclassifications could lead to severe consequences.
For instance, web content writers have faced the career risk due to false-positive classification \cite{gizmodo}. In school education, incorrect detection might ruin students' academic dignity \citep{educatorconsiderations4chatgpt,bloomberg}. 
At the same time, it is extremely difficult, if not impossible, to develop a perfect detector with 100\% accuracy in such real-world scenarios. There remain edge cases in which human-written texts can be misidentified as LLM-generated, and vice versa. Thus, it is crucial to develop a detector that provides interpretable evidence, allowing users to judge how reliably correct the detection results are and identify potential misclassifications \citep{tang2023science,ji2024detectingmachinegeneratedtextsjust}.

% It is extremely difficult to develop a perfect detector with 100\% accuracy in such real-world scenarios. There remain edge cases where human-written texts can be misidentified as LLM-generated and vice versa.
% Thus, it is crucial to develop detectors that provide interpretable evidence, enabling users to assess how reliably correct the predictions are and identify potential misclassifications 

\begin{figure*}[t]
 \begin{center}
 \small
  \centering\includegraphics[width=0.9\textwidth]{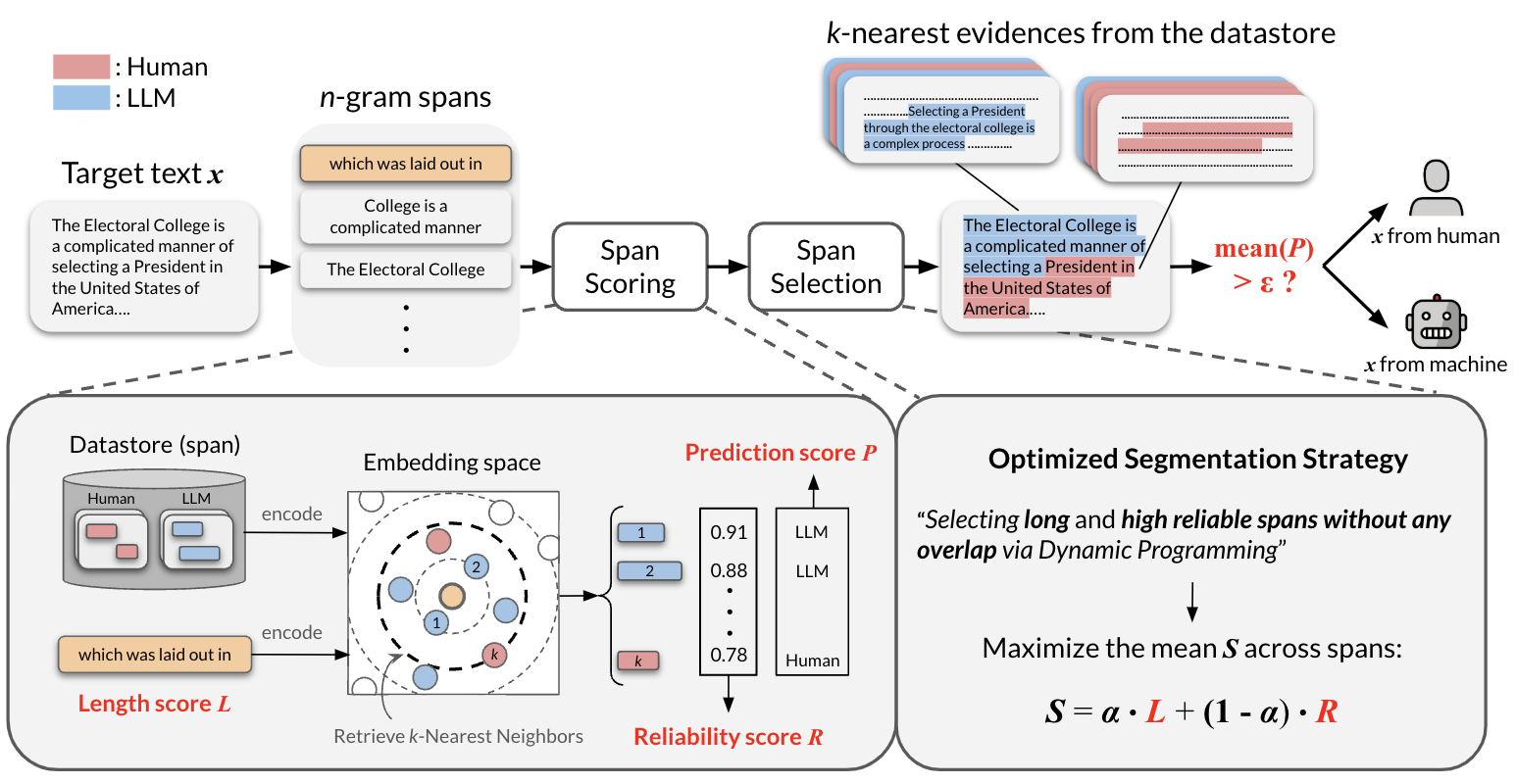}
  \caption{Overview of ExaGPT. It detects the author of a text by examining whether the text shares more similar spans with human-written texts vs. with LLM-generated texts from a datastore.}
  \label{exagpt}
 \end{center}
\end{figure*}

Most detectors lack interpretability of their decisions, outputting only binary prediction labels.
There are few studies on interpretability in detection, and existing approaches provide evidence such as token-level likelihoods \cite{gehrmann2019gltr}, perturbation-based attribution \cite{mitrovic2023chatgpt,wang-etal-2024-m4}, and $n$-gram overlaps between the original text and re-prompted ones \cite{yang2023dnagptdivergentngramanalysis}.
% There are few studies on the interpretability of detection.
% \citet{gehrmann2019gltr} color-highlighted the tokens with high probability under the predicted distribution of LMs. 
% \citet{mitrovic2023chatgpt,wang-etal-2024-m4} showed which part of a text contributed to a decision based on prediction shifts via perturbations to the text.
% \citet{yang2023dnagptdivergentngramanalysis} provided the $n$-gram overlaps between the original text and re-prompted ones by LLMs.
Here, humans intuitively judge text origin by assessing with which source it shares more \textit{similar spans}, including verbatim overlaps and semantically similar spans \cite{maurer06,barron-cedeno-etal-2013-plagiarism}.
However, current detectors are not aligned with the human decision-making process (Figure \ref{human_intuitive_identification}) and fail to yield sufficiently interpretable evidence for users.

% However, humans assess text originality by examining verbatim overlaps or semantically similar spans between the text and existing source texts \cite{MaurerKZ06,barron-cedeno-etal-2013-plagiarism}; prior detectors are not aligned with this decision-making process and still lack interpretability.

Motivated by this gap, we present \textbf{ExaGPT}, an interpretable detection method grounded in the human decision-making process for verifying text origin. ExaGPT makes a prediction by examining whether a text shares more similar spans with human-written or LLM-generated texts from a datastore.
It provides similar span examples that contribute to the decision for each span in the text as evidence.
To present interpretable span-segmented text as a final result, we apply dynamic programming and determine the optimal span break. It balances the long span length and its high frequency (i.e.,~many similar phrases to the span exist in the datastore).
The similarity of the retrieved spans to each span in the target text can help users judge the reliability of the detection result.

To evaluate the interpretability of detection, we conducted a human evaluation of how well people can infer the correctness of the detection from the detector's evidence. We found that
providing similar span examples contributes more effectively to judging the correctness of the detection than existing interpretable methods.
Moreover, extensive experiments in four domains and three generators showed that ExaGPT massively outperforms prior interpretable and powerful detectors by up to +37.0 points of accuracy, even at a constant false positive rate of 1\%. From these results, we observe that ExaGPT achieves high interpretability in its detection result and also high detection performance.\footnote{Code and datasets are available at \url{https://github.com/ryuryukke/ExaGPT}.}

\section{Methodology}
ExaGPT classifies a text based on whether it shares more similar spans with human-written or with LLM-generated texts from a datastore.
As a final result, ExaGPT provides the span-segmented text where each span is accompanied by similar span examples that contribute to the decision.
Figure~\ref{exagpt} illustrates the workflow of ExaGPT, which has two phases: \textbf{Span Scoring} and \textbf{Span Selection}.

In the first phase, we investigate whether each span in the target text shares more similar spans with human-written or LLM-generated texts from a datastore. Meanwhile, we calculate scores for each span, which we use in the second phase (\S\ref{span_scoring}).
% to determine the optimal span segmentation as a final result.

In the second phase, we primarily decide the optimal span segmentation to aid users' understanding of the final result. Specifically, we apply a dynamic programming (DP) algorithm that uses scores from the first phase to find span boundaries, balancing span length and its frequency within the datastore (\S\ref{span_selection}). This yields spans that are both sufficiently long and frequently observed in the datastore. Finally, we detect the target text based on the selected spans and provide similar span examples for each span in the text as interpretable evidence (\S\ref{detection_score}).
We will go into further details below.

\subsection{Span Scoring via \textit{k}-NN Search} 
\label{span_scoring}
Given a target text $x$ to be classified, we define an $n$-gram span in the text $x$ as $x_{i:i+n}$, which is any continuous sequence of $n$ tokens starting in the $i$-th token. For each $n$-gram target span $x_{i:i+n}$, we retrieve the top-$k$ most similar\footnote{We encode the target span, and all spans in the datastore into the same embedding space. We then perform $k$-nearest neighbor ($k$-NN) search based on the cosine similarity of each two span embeddings. See more details in \S\ref{setting_exagpt}.} $n$-gram spans $s_{j}\;(j \in \{1, \dots, k\})$ from the datastore, with each original label and similarity $\{(s_{j}, l_{j}, c_{j})\}_{j=1}^{k}$. Here, $l_{j}$ is \texttt{Human} when the span $s_j$ is part of a human-written text, or \texttt{LLM} when the span $s_j$ is a part of an LLM-generated text. $c_{j}$ is the similarity between the target span $x_{i:i+n}$ and each retrieved span $s_{j}$.

Consequently, we calculate the following metrics for each target span $x_{i:i+n}$: \emph{length score} $L$, \emph{reliability score} $R$, and \emph{prediction score} $P$. 
The length score $L$ is the number of tokens in the target span:
\begin{equation}
L(x_{i:i+n}) = n.
\end{equation}
The reliability score $R$ is the mean similarity $c_{j}$ between the target span and each retrieved span: 
\begin{equation}
R(x_{i:i+n}) = \frac{\sum_{j=1}^{k}c_j}{k}.
\end{equation}
The reliability score $R$ indicates how many similar spans exist in the datastore for the target span. The prediction score $P$ is a ratio of \texttt{LLM} label in the original labels $l_{j}$ of the retrieved spans:
\begin{equation}
P(x_{i:i+n}) = \frac{\sum_{j=1}^{k}\mathbbm{1}(l_j = \text{\texttt{LLM}})}{k} .
\end{equation}
The prediction score $P$ indicates whether the target span shares more similar spans with human-written texts or with LLM-generated texts in the datastore.

\subsection{Span Selection via Dynamic Programming}
\label{span_selection}
In this phase, we select spans $T = [t_1, \dots, t_H]$ in the target text $x$, so that the text is segmented without overlaps as a final result:

\begin{algorithm}[t]
\caption{Span Segmentation Optimization}
\label{dp_algo}
{\linespread{1.09}\selectfont
\begin{algorithmic}
    \STATE {\bfseries Input:} Target text $x$; Length of target text $m$; Length score $L$; Reliability score $R$; Maximum length of \textit{n}-gram span $N$; Hyper-parameter $\alpha$
    \STATE {\bfseries Output:} List of selected \textit{n}-grams $T$
    \STATE ${\rm dp}[0, \ldots, m-1] \leftarrow [([0], {\rm None})] * m$ 
    \FOR{$i=1$ {\bfseries to} $m$}
        \FOR{$j={\rm min}(i - N, 0)$ {\bfseries to} $i$}
        \STATE $l, r \leftarrow  L^{\rm std}(x_{j:i}), R^{\rm std}(x_{j:i})$
        % \STATE Compute 
        % \STATE $l_{\rm std}, r_{\rm std} \leftarrow  {\rm standardize}(l), {\rm standardize}(r)$
        \STATE $scores \leftarrow {\rm dp}[j][0] + [\alpha l + (1-\alpha) r]$
        \STATE $s_{\rm cand} \leftarrow {\rm average}(scores)$
        \IF{${\rm average}({\rm dp}[i][0]) < s_{\rm cand}$}
            \STATE ${\rm dp}[i] \leftarrow (scores, j)$
    \ENDIF
        \ENDFOR
    \ENDFOR
    \STATE Traverse ${\rm dp}$ backward and collect span breaks
    \STATE \textbf{return} List of selected \textit{n}-grams $T$
\end{algorithmic}
}
\end{algorithm}

\begin{equation}
\begin{aligned}
& x = t_1 \oplus t_2 \oplus \dots \oplus t_H, \\
& t_i \cap t_j = \varnothing \quad (i, j \in \{1,\ldots,H\}, i \neq j).
\end{aligned}
\end{equation}
% Here, when humans measure text originality, they focus on whether there are long and similar phrases between the text and other texts.
To facilitate users' understanding of the final result, we optimize the span segmentation that includes longer and more similar spans with ones from the datastore.
Algorithm \ref{dp_algo} describes our dynamic programming strategy to find the best span break. Formally, we select spans $T$ to maximize the score $S$ across the spans in the target text:
\begin{equation}
\label{alpha_formulation}
S(T) = \frac{\sum_{h=1}^{H} \{ \alpha L^{\rm std}(t_h) + (1-\alpha)R^{\rm std}(t_h)\}}{H} .
\end{equation}
Here, $L^{\rm std}(t_h)$ and $R^{\rm std}(t_h)$ are the normalized\footnote{To align the scales of the length score and the reliability score, each score is normalized using the mean and the variance in the validation split of our dataset.} versions of the length score $L$ and the reliability score $R$ of the span $t_h$, respectively. $\alpha$ is an interpolation coefficient ranging from 0.0 to 1.0. $\alpha$ determines the relative contribution of the length score and the reliability score to the span segmentation.

\subsection{Overall Detection with Evidence}
\label{detection_score}
% $[(t_1, P(t_1)), \dots, (t_H, P(t_H))]$
Given a sequence of the selected spans $T$ each with a prediction score for the target text $x$, ExaGPT identifies a text based on the mean prediction score:
\begin{equation}
P_{\rm overall} = \frac{\sum_{h=1}^{H}P(t_h)}{H}.
\end{equation}
ExaGPT classifies a text as \texttt{LLM} if $P_{\rm overall}$ exceeds a detection threshold $\epsilon$, and otherwise as \texttt{Human}.
As evidence of the decision, ExaGPT provides retrieved top-\textit{k} similar spans for each span in the text: 
\begin{equation}
E = [(t_{h}, [s_{h}^{1}, \ldots, s_{h}^{k}])]_{h=1}^{H} .
\end{equation}
The similarity of the retrieved spans to each span in the target text can help users judge how reliably correct the detection result is.

\section{Experiments and Results}
\label{experiments}
% In this section, we explore the interpretability and detection performance of the ExaGPT.

\subsection{Overall Setup}
\label{setup}

\paragraph{Metrics.}
To assess detection performance, we use the AUROC score, which is widely used in detection studies. However, it is only useful to observe the overall behavior of a detector through all possible thresholds. In practice, it is critical to minimize the false positive classification, i.e.,~wrongly identifying human-written texts as LLM-generated. We thus report detection accuracy with a threshold by fixing the false-positive rate (FPR) at 1\%, a common evaluation metric in recent robustness studies \cite{krishna2023paraphrasing,hans2024spotting,dugan-etal-2024-raid}.

\paragraph{Datasets.}
We use the M4 dataset \cite{wang-etal-2024-m4}, a large-scale detection benchmark comprising pairs of human-written and LLM-generated texts across multiple languages, domains, and generators. Our experiments use the English subset, including 3,000 pairs of human-written and LLM-generated texts from each combination of four domains: Wikipedia, Reddit, WikiHow, and arXiv, and three generators: ChatGPT, GPT-4 as closed-source LLMs, and Dolly-v2 \cite{dolly} as open-source LLMs. For each combination, we split the dataset into three parts: train/valid/test with 2,000/500/500 pairs, respectively.

\paragraph{Baselines.}
We compare ExaGPT to three strong and interpretable detectors, including RoBERTa with SHAP \cite{mitrovic2023chatgpt}, LR-GLTR \cite{wang-etal-2024-m4}, and DNA-GPT \cite{yang2023dnagptdivergentngramanalysis}, whose details are provided in \S\ref{interpretable_detectors}.

The first one is a supervised classifier based on RoBERTa \cite{liu2019roberta}, which we fine-tune for detection on our train split. 
Similarly, we train the LR-GLTR detector on our train split with selected and hand-crafted GLTR features \cite{gehrmann2019gltr}, following \citet{wang-etal-2024-m4}. The hyperparameter settings for training both RoBERTa and LR-GLTR are aligned with those of \citet{wang-etal-2024-m4}. Further configurations of the baselines are in Appendix~\ref{configurations}.

\paragraph{ExaGPT.}
\label{setting_exagpt}
In the span scoring phase, ExaGPT uses our train split as the datastore for each combination of domains and generators. 
We consider the $n$-gram size to be from 1 to 20 across the entire dataset.
We embed the target span and all spans in the datastore into the same vector space using BERT\footnote{\url{https://huggingface.co/google-bert/bert-large-uncased}}, a standard embedding model. For a span embedding, we feed a text into BERT and take the mean hidden states\footnote{We select the second layer where the \textit{k}-NN spans are similar to the target span well-balanced lexically and semantically, enhancing its interpretability in our pilot study.} of the tokens within the span.
We retrieve the top-\textit{k} (=10)\footnote{We choose the value of \textit{k} so that ExaGPT shows favorable detection performance over smaller values in our pilot study and does not reduce the interpretability. Since ExaGPT presents retrieved spans as evidence, keeping \textit{k} small helps users assess detection correctness based on a manageable amount of information.} most similar spans from the datastore for each target span via $k$-NN search using the FAISS \cite{faiss_paper}.

% In the span selection phase, we select the optimal $\alpha$ from a predefined set of values in the range $[0,1]$ with a step size of 0.125, where ExaGPT exhibits the best detection performance in our validation split. The selected $\alpha$ is then fixed and used consistently throughout all subsequent evaluations, including both interpretability and detection performance of ExaGPT.

% In the span selection phase, we select the optimal $\alpha$ from a predefined set of values in the range $[0,1]$ with a step size of 0.125, based on detection performance on the validation split. 
% Specifically, we choose the value that yields the best performance of ExaGPT on this split. 
% The selected $\alpha$ is then fixed and used consistently throughout all subsequent evaluations, including both interpretability and detection performance.

In the span selection phase, we select the optimal $\alpha$ from values between 0.0 and 1.0 at 0.125 intervals, where ExaGPT exhibits the best detection performance in our validation split. The $\alpha$ is constant throughout our evaluation of the interpretability
and the detection performance of ExaGPT.

\paragraph{Human Evaluation on Interpretability.}
We assess the interpretability of detectors through human evaluation, since it is crucial that detectors provide evidence enabling users to judge the reliability of detection results. Therefore, we first design a human evaluation to test whether the provided evidence \textit{actually helps} users determine whether a detection is correct, a practical aspect overlooked in prior work. Participants are shown the detection evidence and result, and are asked to judge their correctness. Accordingly, our interpretability metric is defined as the accuracy of these human judgments on detection correctness.

\begin{figure}[t]
 \begin{center}
  \centering\includegraphics[width=\columnwidth]{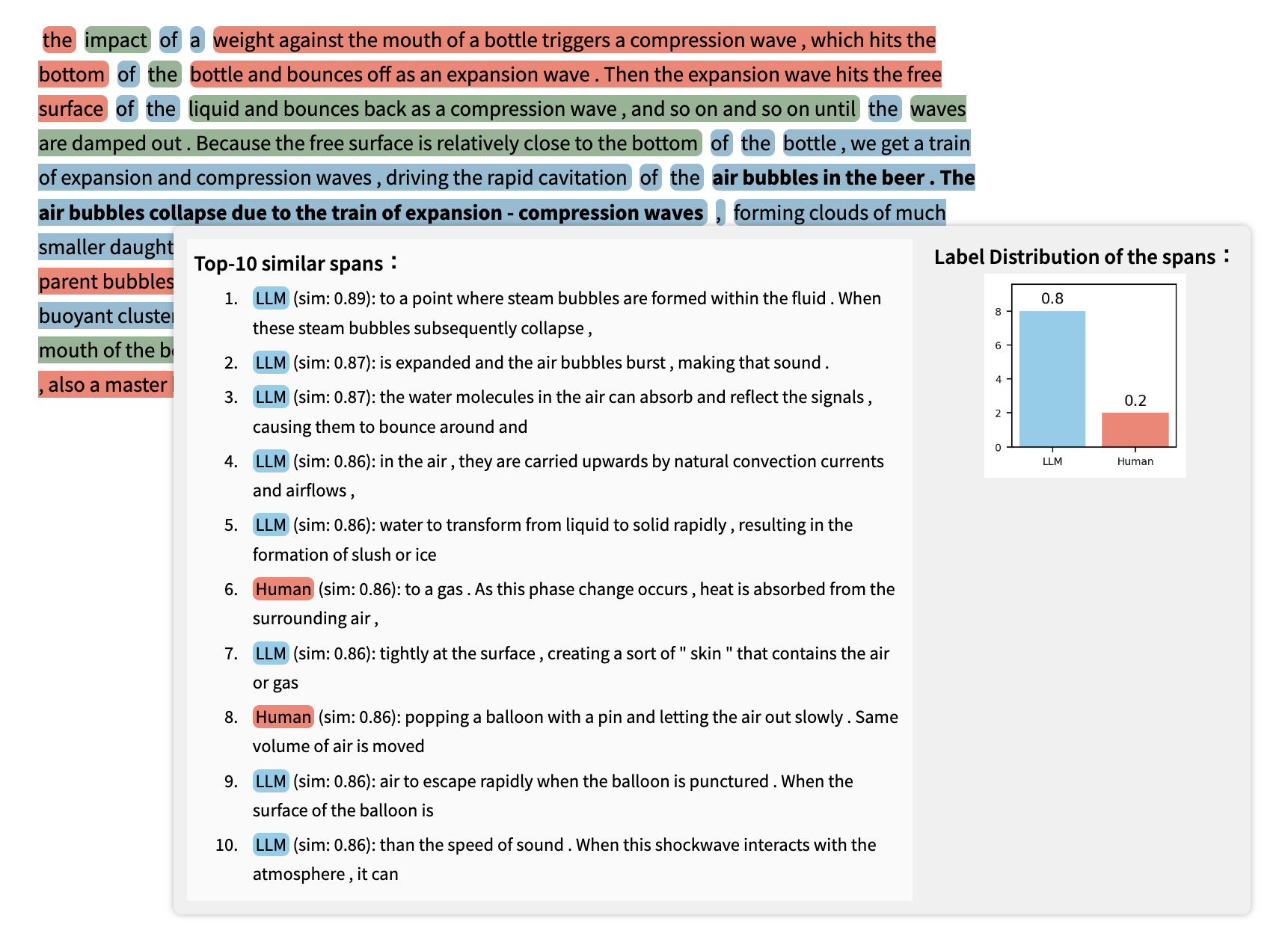}
  \caption{User interface of ExaGPT. Hovering over a text span displays the tooltip about the retrieved similar spans each with the similarity to the span and the original label distribution.}
  \label{exagpt_demo}
 \end{center}
\end{figure}

For each detector, we evaluate 96 samples\footnote{The 96 samples for each detector consist of two samples (one correct and one incorrect) across four domains and three generators, distributed among four participants.} from our test split across all combinations of domains and generators, with an equal ratio of correct and incorrect detections. Four annotators (one MSc student, one PhD student, and two NLP researchers)\footnote{They are not authors of this paper. Although all annotators have NLP backgrounds, they vary in research experience and language proficiency (including both native and non-native English speakers), providing diverse perspectives.} were provided with different samples.
Figure \ref{exagpt_demo} shows the user interface of ExaGPT in the evaluation. 
Spans are highlighted\footnote{ExaGPT performs the overall detection rather than detecting each span individually. However, for better readability, each span is color-highlighted on its prediction score.} in red, green, and blue for which prediction score $P$ is lower than 0.5 (human-written), equal to 0.5 (neither), and higher than 0.5 (LLM-generated). Participants judge the correctness of the detection by mainly examining similar span examples. Details of baseline evidence are provided in Appendix~\ref{evidence_baselines}.

% \paragraph{Justification of sample size.}
% ssss
\subsection{Results}
\label{results}
\paragraph{Detection Interpretability.}
Table \ref{detection_interpretability_result} presents the differences in the accuracy of human judgments on detection correctness based on the provided evidence across baseline detectors and ExaGPT. The accuracy of human judgments on ExaGPT is relatively higher compared to baseline detectors by up to +13.6 points. This indicates that ExaGPT provides more interpretable evidence than other baselines, helping humans judge the correctness of detections more effectively.

Specifically, DNA-GPT also provides $n$-gram span overlaps between the target text and re-generated LLM texts from the truncated part as evidence.
The comparison of the human evaluation score between DNA-GPT and ExaGPT suggests that providing not only simple overlaps but also semantically similar spans contributes to better interpretability.
We further investigate how the similarity between the target span and retrieved spans correlates with the correctness of the detection of ExaGPT in \S\ref{what_makes_exagpt_interpretable}.

\begin{table}[t]
\centering
\small
\setlength{\tabcolsep}{5pt} % Default value: 6pt
\renewcommand{\arraystretch}{1} % Default value: 1
\begin{tabular}{lc}
\toprule
\multicolumn{1}{c}{\textbf{Detector}} & \textbf{ACC. of Human Judgements (\%) $\uparrow$} \\
\midrule
RoBERTa & 47.9 \\
LR-GLTR & 57.3 \\
DNA-GPT & 53.1 \\
ExaGPT & \textbf{61.5} \\
\bottomrule
\end{tabular}
\caption{Comparison of the accuracy (ACC.) of human judgments on the correctness of detections based on evidence across baseline detectors and ExaGPT. Higher accuracy implies that the detector provides more interpretable evidence to users.}
\label{detection_interpretability_result}
\end{table}

\paragraph{Detection Performance.}
Table \ref{detection_performance_result} presents the differences in detection performance between baseline detectors and ExaGPT across domains and generators. The detection performance includes AUROC and the accuracy at 1\% FPR. 
Overall, ExaGPT consistently demonstrates detection performance on par with or better than baselines, including supervised classifiers. Specifically, on accuracy at 1\% FPR, ExaGPT achieves the best average detection performance on all three generators, outperforming baselines by a large margin of up to +37.0 points. This suggests that ExaGPT is the most effective detector in practical scenarios, where we need to minimize the false positives.

% \paragraph{Remark.} From the experiments on interpretability and classification performance of detectors, we observe that ExaGPT achieves both superior interpretability of the detection and exceptional detection performance compared to previous interpretable detectors.

% \paragraph{Summary.} 
% Experiments demonstrate that ExaGPT achieves both superior interpretability and stronger detection performance compared to prior interpretable detectors.

% ExaGPT achieved both superior interpretability of the detection and exceptional detection performance compared to previous interpretable detectors.

\begin{table*}[h]
\centering
\small
\fontsize{8.5pt}{9.5pt}\selectfont
\setlength{\tabcolsep}{5pt} % Default value: 6pt
\renewcommand{\arraystretch}{0.95} % Default value: 1
\begin{tabular}{clcccccccccc}
\toprule
 \multirow{2.5}{*}{\textbf{Generator}} & \multicolumn{1}{c}{\multirow{2.5}{*}{\textbf{Detector}}} & \multicolumn{2}{c}{\textbf{Wikipedia}} & \multicolumn{2}{c}{\textbf{Reddit}} & \multicolumn{2}{c}{\textbf{WikiHow}} & \multicolumn{2}{c}{\textbf{arXiv}} & \multicolumn{2}{c}{\textbf{Average}} \\
\cmidrule(lr){3-4} \cmidrule(lr){5-6} \cmidrule(lr){7-8} \cmidrule(lr){9-10} \cmidrule(lr){11-12} 
 & & AUROC & ACC. & AUROC & ACC. & AUROC & ACC. & AUROC & ACC. & AUROC & ACC. \\
\midrule
\multirow{4}{*}{ChatGPT} & RoBERTa & \textbf{100.0} & \underline{77.1} & \textbf{99.8} & 61.0 & \textbf{100.0} & 50.0 & \textbf{100.0} & 87.3 & \textbf{100.0} & 68.9 \\
& LR-GLTR & 95.0 & 60.0 & \underline{99.4} & \textbf{94.0} & 97.5 & 85.8 & \underline{99.8} & \textbf{97.7} & 97.9 & \underline{84.4}\\
& DNA-GPT & 84.8 & 49.4 & 92.3 & 62.9 & 99.4 & \underline{93.5} & 89.0 & 59.9 & 91.4 & 66.4\\
& ExaGPT & \underline{98.6} & \textbf{92.3} & 98.9 & \underline{86.6} & \underline{99.5} & \textbf{96.0} & 99.6 & \underline{95.8} & \underline{99.2} & \textbf{92.7} \\
\midrule
\multirow{4}{*}{GPT-4} & RoBERTa & \textbf{100.0} & \textbf{87.8} & \textbf{100.0} & 66.4 & \textbf{100.0} & 77.4 & \textbf{100.0} & 68.6 & \textbf{100.0} & 75.1 \\
& LR-GLTR & 97.8 & 85.7 & \underline{99.6} & \textbf{97.2} & 94.8 & \underline{77.8} & \textbf{100.0} & \underline{98.5} & 98.1 & \underline{89.8}\\
& DNA-GPT & 40.3 & 48.1 & 71.9 & 68.6 & 44.6 & 49.9 & 72.2 & 54.4 & 57.3 & 55.3\\
& ExaGPT & \underline{98.3} & \underline{87.3} & 99.3 & \underline{91.1} & \underline{98.8} & \textbf{92.2} & \underline{99.7} & \textbf{98.7} & \underline{99.0} & \textbf{92.3}\\
\midrule
\multirow{4}{*}{Dolly-v2} & RoBERTa & \textbf{100.0} & \underline{61.8} & \textbf{100.0} & 50.0 & \textbf{100.0} & 70.8 & \textbf{100.0} & \textbf{82.8} & \textbf{100.0} & 66.4 \\
& LR-GLTR & 79.7 & 57.7 & 95.3 & \textbf{79.0} & 72.4 & 55.0 & \underline{93.7} & \underline{78.2} & 85.3 & \underline{67.5} \\
& DNA-GPT & 68.0 & 61.5 & 67.5 & 66.1 & 87.7 & \textbf{82.3} & 64.9 & 57.7 & 72.0 & 66.9 \\
& ExaGPT & \underline{85.8} & \textbf{63.8} & \underline{96.2} & \underline{76.6} & \underline{94.3} & \underline{75.6} & 85.2 & 67.3 & \underline{90.4} & \textbf{70.8} \\
\bottomrule
\end{tabular}
\caption{Comparison of detection performances of ExaGPT and baseline detectors on texts from various domains and generators. \emph{ACC.} indicates the detection accuracy at 1\% FPR. \textbf{Bold} and \underline{Underline} indicate the best and runner-up performance for each combination of domains and generators.}
\label{detection_performance_result}
\end{table*}

\section{Analysis}
% In this section, we explore the reason behind the high interpretability of ExaGPT. We specifically focus on the span length and the semantic similarity between each target span and retrieved spans, which are prioritized in the span selection.
% Furthermore, we investigate the robustness of the detection performance of ExaGPT with respect to the interpolation coefficient $\alpha$ and the size of the datastore, respectively. 
% we explore the reason behind the high interpretability of ExaGPT, specifically in terms of the proximity between the target span and retrieved \textit{k}-NN spans.

\paragraph{What Makes ExaGPT Interpretable.}
\label{what_makes_exagpt_interpretable}
Our human evaluations demonstrate that ExaGPT provides highly interpretable evidence compared to prior detectors. To explore the reason for this, we investigate the difference in the characteristics of the selected spans as a final output between correct and incorrect predictions by ExaGPT.
We focus on span length and mean similarity between each target span and the retrieved spans (reliability score $R$), which are prioritized in the span selection. We randomly select 1,000 correct and 1,000 incorrect ExaGPT predictions on our test splits across all combinations of domains and generators.

\begin{table*}[t]
\centering
\small
\setlength{\tabcolsep}{6pt} % Default value: 6pt
\renewcommand{\arraystretch}{1.14} % Default value: 1
\begin{tabular}{c|c|l}
\toprule
\textbf{Target Span} & \llmmarker{\texttt{LLM}} & \textit{published in 1993. The novel tells the story of a young Jewish slave, Hadassah,} \\
\midrule
\multirow{10}{*}{\textbf{\textit{k}-NN Spans}} & \llmmarker{\texttt{LLM (0.92)}} & \textit{and was first published in 1936. The book tells the story of three orphaned sisters,} \\
 & \llmmarker{\texttt{LLM (0.92)}} & \textit{published in 2012. The novel revolves around the story of a young woman} \\
 & \llmmarker{\texttt{LLM (0.90)}} & a\textit{nd published in 2010. The novel tells the story of Michael Beard, a}\\
 & \llmmarker{\texttt{LLM (0.90)}} & \textit{ling of the biblical book, Song of Solomon, and is considered one of the}\\
 & \llmmarker{\texttt{LLM (0.90)}} & \textit{man and published in 1963. The book was later adapted into a Disney film of the}\\
 & \llmmarker{\texttt{LLM (0.90)}} & \textit{. The film tells the story of a young}\\
 & \humanmarker{\texttt{Human (0.89)}} & \textit{the Xanth series. It is the second book of a trilogy beginning with Vale of the}\\
 & \llmmarker{\texttt{LLM (0.89)}} & \textit{published in 1959. The novel is set in the Arctic region and follows the story of Dr.} \\
 & \humanmarker{\texttt{Human (0.89)}} & \textit{. It is the third novel in the Dahak trilogy, after the de} \\
 & \llmmarker{\texttt{LLM (0.89)}} & \textit{for his semi-autobiographical novel, ``The Watch that Ends the Night''. Born in} \\
\bottomrule
\end{tabular}
\caption{Examples of $k$-NN spans for a target span retrieved by ExaGPT. The colored part represents the original label for each span (\texttt{LLM} in \textcolor{blue}{blue} and \texttt{Human} in \textcolor{red}{red}, respectively). In the part of $k$-NN spans, the similarity between the target span and each $k$-NN span is added.}
\label{knn_spans}
\end{table*}

\begin{figure}[t]
 \begin{center}
  \centering\includegraphics[width=0.75\columnwidth]{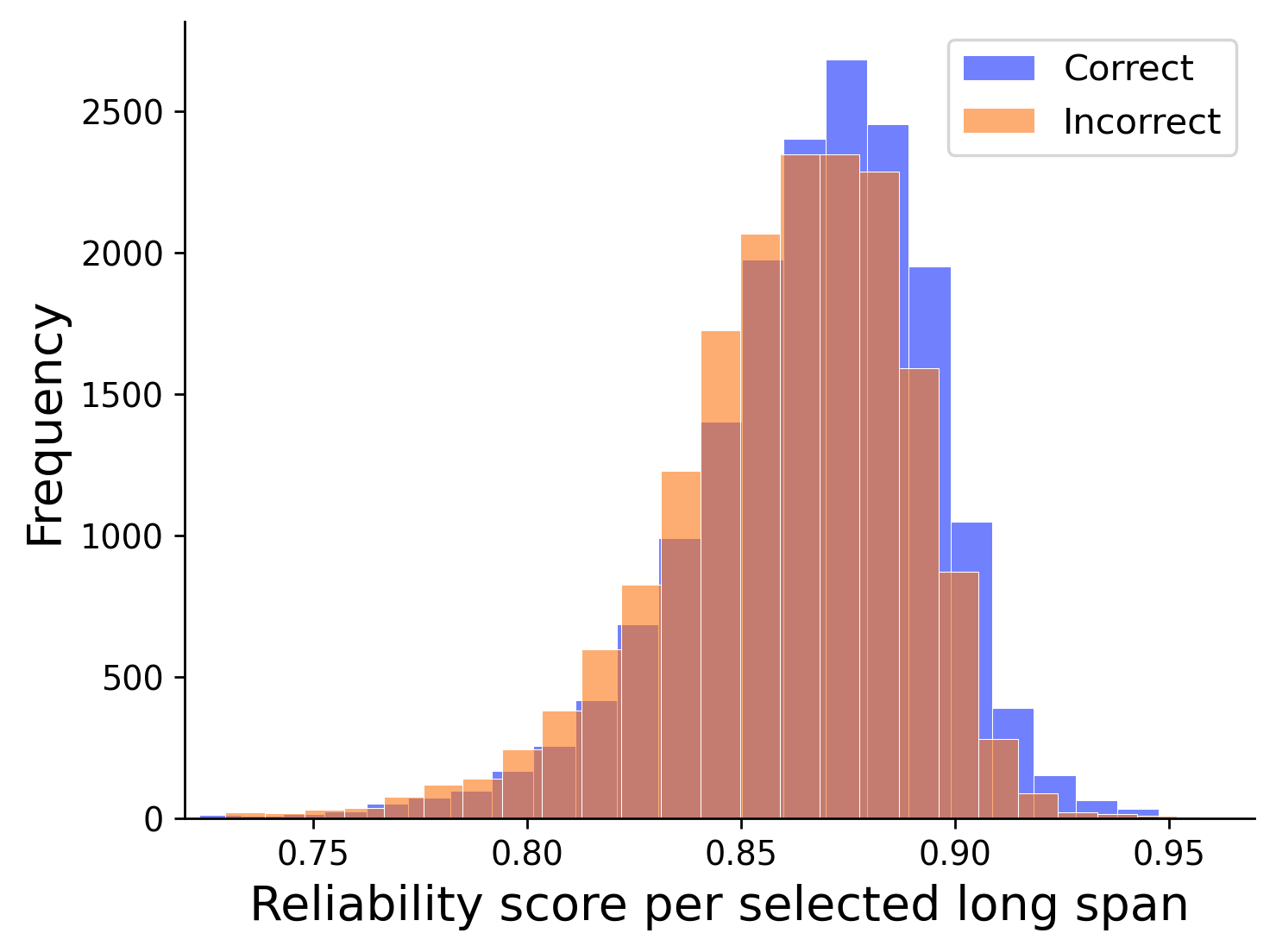}
  \caption{Reliability score distributions of long spans (\textit{n} $\geq$ 10) in correct and incorrect samples of ExaGPT.}
  \label{reliable_score_variation}
 \end{center}
\end{figure}

% \begin{figure}[t]
%  \begin{center}
%   \centering\includegraphics[width=0.77\columnwidth]{figures/alpha_shift_chatgpt.png}
%   \caption{Impact of $\alpha$ on the detection performance of ExaGPT, using ChatGPT as a generator.}
%   \label{alpha_effect}
%  \end{center}
% \end{figure}

Figure \ref{reliable_score_variation} presents the reliability score distributions for long spans (\textit{n} $\geq$ 10) in correct and incorrect samples. 
A rightward shift indicates that correct samples of ExaGPT include more long spans with higher reliability scores than incorrect ones. 
This suggests that offering long spans with high reliability scores helps users judge the correctness of the detections.

Table~\ref{knn_spans} presents examples of long spans (\textit{n} $=$ 19) with high reliability scores for a target span retrieved by ExaGPT.
We can see that the retrieved spans are well-balanced between lexical and semantic similarity to the target span. 
% Due to space limitations, the full table is provided in Appendix~\ref{analysis_details}.

\paragraph{Impact of $\boldsymbol{\alpha}$.}
\label{alpha}
In our experiments, we determine the optimal coefficient $\alpha$ for ExaGPT (as used in Eq.~\ref{alpha_formulation}) based on the best detection performance on the validation split.
To examine the robustness of ExaGPT to the choice of $\alpha$, we analyze how detection performance varies as $\alpha$ changes.
\begin{figure}[t]
 \begin{center}
  \centering\includegraphics[width=0.79\columnwidth]{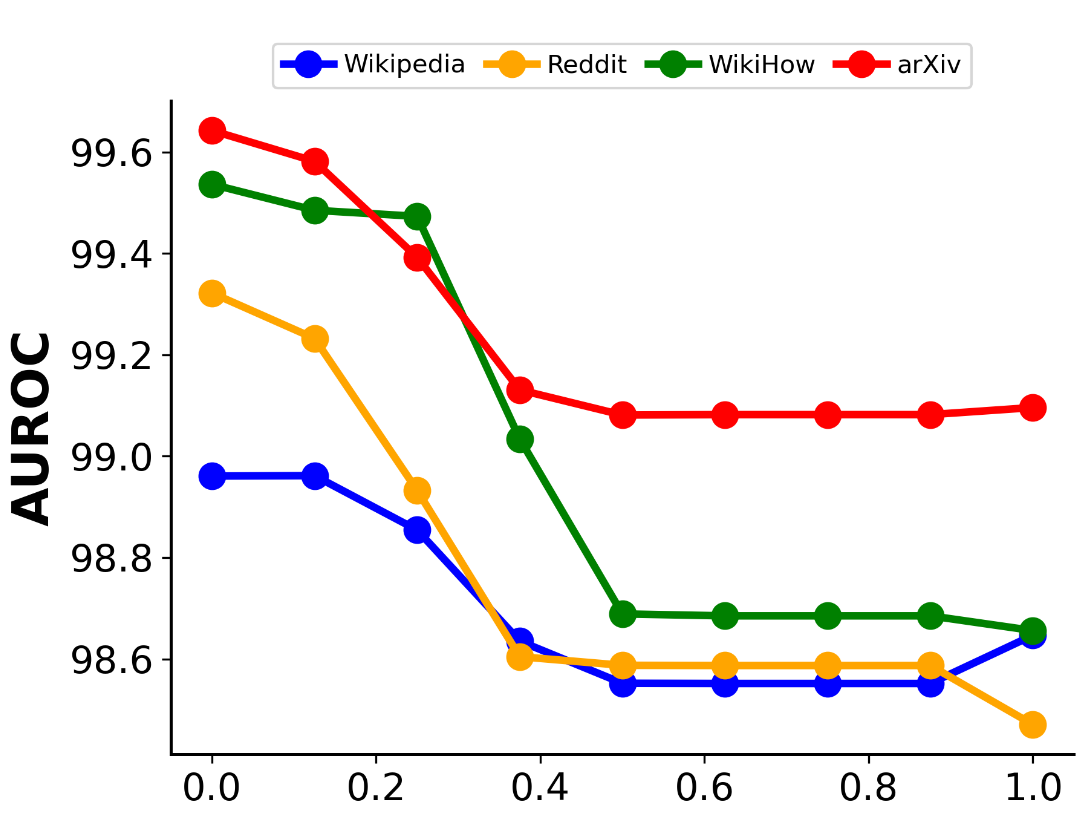}
  \caption{Impact of $\alpha$ on the detection performance of ExaGPT, using ChatGPT as a generator.}
  \label{alpha_effect}
 \end{center}
\end{figure}

\begin{figure}[t]
 \begin{center}
  \centering\includegraphics[width=0.79\columnwidth]{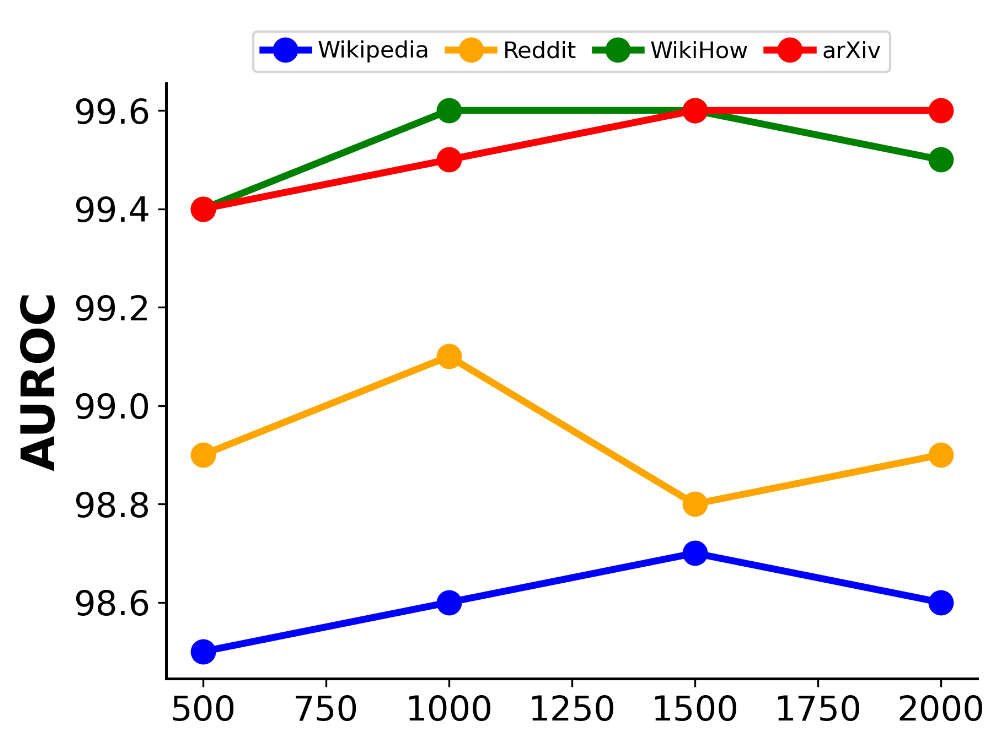}
  \caption{Impact of the datastore size on the detection performance of ExaGPT, using ChatGPT as a generator.}
  \label{datastore_size_effect_chatgpt}
 \end{center}
\end{figure}

Figure \ref{alpha_effect} depicts the relationship between $\alpha$ and the detection performance of ExaGPT, evaluated on ChatGPT-generated text across four domains.
$\alpha$ ranges from 0.0 to 1.0 in increments of 0.125. We observe that larger values of $\alpha$ generally lead to lower detection performance.
This suggests that placing greater weight on the reliability score (i.e., selecting target spans that are more similar to spans in the datastore) improves detection performance.

Notably, across all four domains, the lowest AUROC is 98.5\%, suggesting that changes in $\alpha$ do not cause a substantial performance drop that would change the ranking of detectors. See Appendix \ref{analysis_details} for consistent trends in all generators.

\paragraph{Impact of Datastore Size.}
\label{datastore}
In our evaluation, ExaGPT uses the train split as the datastore from which it retrieves the top-\textit{k} most similar spans for each span in a target text.
To study the robustness of ExaGPT to datastore size, we analyze how detection performance varies as the datastore size changes. The train split contains 2,000 pairs of human-written and LLM-generated texts. We randomly sample \{500, 1,000, 1,500, 2,000\} pairs from the train split as datastores of different sizes.

Figure \ref{datastore_size_effect_chatgpt} shows the relationship between datastore size and the detection performance of ExaGPT across four domains using ChatGPT as the generator.
Overall, ExaGPT remains robust to datastore size, exhibiting only minor performance degradation. Interestingly, ExaGPT with a datastore of 500 pairs performs comparably to using the full 2,000 pairs in terms of AUROC.
See Appendix \ref{analysis_details} for consistent trends in all generators.

\begin{table}[t]
\centering
\Large
\renewcommand{\arraystretch}{1.7}
\resizebox{\columnwidth}{!}{%
\begin{tabular}{ccccccc}
\toprule
 & & \multicolumn{4}{c}{\textbf{Test}} \\
\cmidrule(lr){3-6}
 & & \textbf{Wikipedia} & \textbf{Reddit} & \textbf{WikiHow} & \textbf{arXiv} & \textbf{Average} \\
\midrule
\multirow{5.5}{*}{\rotatebox{90}{\textbf{Train}}}
& Wikipedia & \textbf{98.3 / 87.3} & 91.7 / 68.2 & 54.1 / 53.3 & 89.3 / 60.5 & 83.4 / \underline{67.3} \\
& Reddit    & 90.1 / 60.7 & \textbf{99.3 / 91.1} & 74.6 / 50.6 & 93.0 / 63.9 & \underline{89.3} / 66.6 \\
& WikiHow   & 66.2 / 50.6 & 76.9 / 60.4 & \textbf{98.8 / 92.2} & 64.7 / 51.6 & 76.7 / 63.7 \\
& arXiv     & 73.7 / 50.4 & 86.3 / 56.2 & 57.0 / 51.5 & \textbf{99.7 / 98.7} & 79.2 / 64.2 \\
\cmidrule(lr){2-7}
& ALL & \underline{94.3} / \underline{80.7} & \underline{96.7} / \underline{83.5} & \underline{92.9} / \underline{73.4} & \underline{99.5} / \underline{96.7} & \textbf{95.9} / \textbf{83.6} \\
\bottomrule
\end{tabular}
}
\caption{Cross-domain detection with GPT-4 as the generator. The scores are AUROC / Acc@FPR=1\%.}
\label{cross-domain}
\end{table}

\begin{table}[t]
\centering
\renewcommand{\arraystretch}{1.3}
\resizebox{\columnwidth}{!}{%
\begin{tabular}{llcccc}
\toprule
 & & \multicolumn{3}{c}{\textbf{Test}} \\
\cmidrule(lr){3-5}
 & & \textbf{ChatGPT} & \textbf{GPT-4} & \textbf{Dolly} & \textbf{Average} \\
\midrule
& ChatGPT & \textbf{99.6 / 95.8} & 98.2 / 84.5 & 63.2 / 50.3 & 87.0 / \underline{76.9} \\
\multirow{2}{*}{\rotatebox{90}{\textbf{Train}}}
& GPT-4   & 94.6 / 66.6 & \textbf{99.7 / 98.7} & 61.8 / 51.5 & 85.4 / 72.3 \\
& Dolly   & 93.0 / 69.9 & 89.9 / 65.5 & \textbf{85.2 / 67.3} & \underline{89.4} / 67.6 \\
\cmidrule(lr){2-6}
& ALL & \underline{98.9} / \underline{91.4} & \underline{99.3} / \underline{95.6} & \underline{76.4} / \underline{52.9} & \textbf{91.5} / \textbf{80.0} \\
\bottomrule
\end{tabular}
}
\caption{Cross-generator detection with arXiv as the domain. The scores are AUROC / Acc@FPR=1\%.}
\label{cross-generator}
\end{table}

\begin{table}[H]
\centering
\renewcommand{\arraystretch}{1.4}
\resizebox{\columnwidth}{!}{%
\begin{tabular}{lccccc}
\toprule
\textbf{Detector} & \textbf{Wikipedia} & \textbf{Reddit} & \textbf{WikiHow} & \textbf{arXiv} & \textbf{Average} \\
\midrule
LR-GLTR
& 89.4 / 60.2 
& 97.0 / \textbf{76.8} 
& 89.5 / 58.0 
& \textbf{99.6 / 96.7} 
& 93.9 / 72.9 \\
ExaGPT
& \textbf{98.0 / 86.5} 
& \textbf{97.2} / 69.4 
& \textbf{91.1 / 73.8} 
& 97.7 / 76.4 
& \textbf{96.0 / 76.5} \\
\bottomrule
\end{tabular}
}
\caption{Performance on paraphrased text. The scores are AUROC / Acc@FPR=1\%.}
\label{paraphrasing}
\end{table}

% \begin{table}[H]
% \centering
% \renewcommand{\arraystretch}{1.4}
% \resizebox{\columnwidth}{!}{%
% \begin{tabular}{lccccc}
% \toprule
% \textbf{Detector} & \textbf{Wikipedia} & \textbf{Reddit} & \textbf{WikiHow} & \textbf{arXiv} & \textbf{Average} \\
% \midrule
% LR-GLTR
% & 89.4 / 60.2 
% & 97.0 / \textbf{76.8} 
% & 89.5 / 58.0 
% & \textbf{99.6 / 96.7} 
% & 93.9 / 72.9 \\
% ExaGPT
% & \textbf{98.0 / 86.5} 
% & \textbf{97.2} / 69.4 
% & \textbf{91.1 / 73.8} 
% & 97.7 / 76.4 
% & \textbf{96.0 / 76.5} \\
% \bottomrule
% \end{tabular}
% }
% \caption{Performance on paraphrased text. The scores are AUROC / Acc@FPR=1\%.}
% \label{paraphrasing}
% \end{table}

\paragraph{Unknown Domain or Generator.}
While our primary goal is to improve interpretability, we also perform cross-domain and cross-generator experiments to examine how ExaGPT can be leveraged in more realistic settings.
Tables~\ref{cross-domain} and \ref{cross-generator} report the results, where ``ALL'' denotes a datastore constructed from all domains or generators, with samples drawn uniformly to match the size of the single-source setting.

We observe that when the domain or generator in the datastore is different from the target model or generator, detection performance is reduced. 
However, using multiple sources in the datastore, instead of relying on only a single source, is considered more realistic in practice.
Our results confirm that in such a setting (``ALL''), ExaGPT consistently maintains strong performance across multiple domains and generators.
The results further suggest which domains or generators are effective for generalization. For instance, including Reddit mitigates AUROC drops across domains, whereas including ChatGPT helps maintain high detection performance against GPT-4.

\paragraph{Paraphrased Text.}
We also investigate the robustness of ExaGPT against paraphrased text.
Following \citet{krishna2023paraphrasing}, we utilize DIPPER, an 11B document-level paraphraser, to rewrite machine-generated text. Table \ref{paraphrasing} reports results on all domains using ChatGPT as the generator. Among strong interpretable detectors, LR-GLTR was the runner-up in our original in-domain evaluation (\S\ref{results}). Even under paraphrasing, ExaGPT maintains moderately high detection performance and generally outperforms LR-GLTR.

\begin{figure}[t]
 \begin{center}
  \centering\includegraphics[width=0.98\columnwidth]{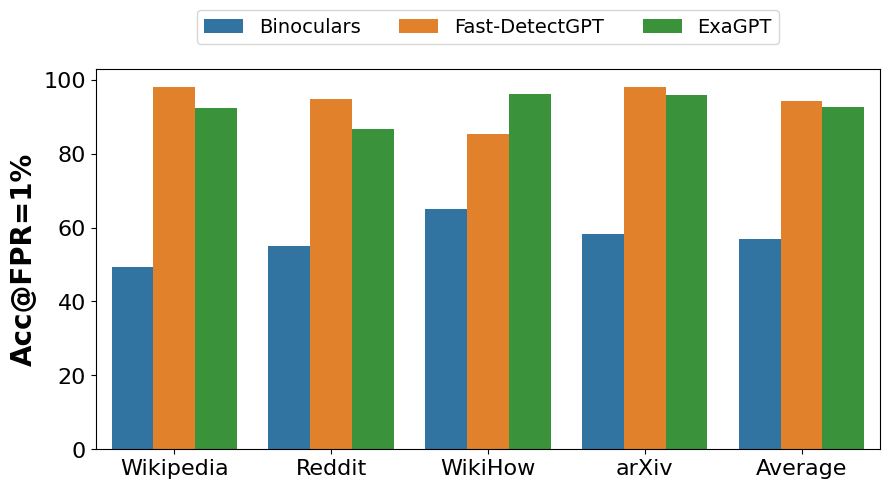}
  \caption{Performance comparison of ExaGPT with state-of-the-art non-interpretable detectors on Acc@FPR=1\%, using ChatGPT as a generator.}
  \label{sota}
 \end{center}
\end{figure}

\paragraph{Comparison with state-of-the-art Detectors.}
To understand the trade-off between interpretability and detection performance, we compare ExaGPT with state-of-the-art non-interpretable detectors.
Figure \ref{sota} reports results for Binoculars \citep{hans2024spotting} and Fast-DetectGPT \citep{bao2024fastdetectgptefficientzeroshotdetection}, which are strong metrics-based detectors. The evaluation is conducted across all domains with ChatGPT.
Notably, despite being interpretable, ExaGPT achieves detection performance on par with or even better than these state-of-the-art detectors.

\paragraph{Inference Cost.}

We evaluate the inference cost of ExaGPT, focusing on the \textit{k}-NN search, which is the primary bottleneck. We measure its latency and GPU memory usage and further evaluate FAISS-based approximate search with IVFPQ to improve inference speed and reduce resource usage.

Table~\ref{inference} shows results on WikiHow with ChatGPT. Reducing the datastore from 2,000 to 500 pairs lowers memory by 66\% and latency by 60\% with almost no performance loss. Using FAISS-based approximation further reduces memory by 87\% and latency by over 90\%, while the performance drop is still moderate. These results demonstrate that ExaGPT can be deployed efficiently under practical computational budgets.

\begin{table}[t]
\centering
\normalsize
\renewcommand{\arraystretch}{1.3}
\setlength{\tabcolsep}{4.5pt}

\resizebox{\columnwidth}{!}{%
\begin{tabular}{lcccc}
\toprule
\raisebox{0.6ex}{\textbf{Setting}}
& \raisebox{0.6ex}{\textbf{\#Inst.}}
& \shortstack{\textbf{GPU memory}\\\textbf{(GB)}}
& \shortstack{\textbf{Latency}\\\textbf{(sec.)}}
& \raisebox{0.6ex}{\textbf{AUROC}} \\
\midrule
2000 pair
& 36M
& 162.2
& 14.6
& 99.5 \\
500 pair
& 9.1M
& 54.7 (66\%$\downarrow$)
& 5.81 (60\%$\downarrow$)
& 99.4 \\
500 pair + IVFPQ
& 9.1M
& 20.2 (87\%$\downarrow$)
& 1.22 (90\%$\downarrow$)
& 97.8 \\
\bottomrule
\end{tabular}%
}
\caption{Inference cost analysis of ExaGPT. \#Inst. refers to the number of \textit{n}-gram spans in the datastore.}
\label{inference}
\end{table}

\section{Related Work}
\paragraph{Machine-Generated Text Detection.} 
Prior studies can be grouped into three categories: \emph{text watermarking}, \emph{metrics-based}, and \emph{supervised classifiers}.
Text watermarking modifies the decoding process so that selected tokens appear more frequently, enabling detection by checking their ratio in the output \cite{kirchenbauer2023watermark}.
Metrics-based methods measure the probabilistic discrepancy of a text with the model's predicted distribution, using signals such as token log probabilities \cite{gehrmann2019gltr}, token ranks \cite{solaiman2019release,su2023detectllm}, entropy \cite{ent08}, perplexity \cite{bere16,hans2024spotting}, and negative probability curvature  \cite{mitchell2023detectgpt,bao2024fastdetectgptefficientzeroshotdetection}.
Supervised classifiers are models fine-tuned to discern human-written and LLM-generated texts with labels. They vary from probabilistic \cite{ippolito-etal-2020-automatic,crothers2023machinegeneratedtextcomprehensive} to neural methods \cite{uchendu-etal-2020-authorship,guo2023close,emi2024technicalreportpangramaigenerated}. 

% \begin{table}[t]
% \centering
% \Large
% \renewcommand{\arraystretch}{1.5}
% \resizebox{\columnwidth}{!}{%
% \begin{tabular}{lccccc}
% \toprule
% \textbf{Detector} & \textbf{Wikipedia} & \textbf{Reddit} & \textbf{WikiHow} & \textbf{arXiv} & \textbf{Average} \\
% \midrule
% Binoculars 
% & 83.4 / 49.4 
% & 72.4 / 55.0 
% & 80.9 / 65.0 
% & 84.1 / 58.2 
% & 80.2 / 56.9 \\
% F.GPT
% & \textbf{99.6 / 98.1} 
% & \textbf{99.4 / 94.7} 
% & \underline{95.8} / \underline{85.4} 
% & \textbf{100.0 / 98.1} 
% & \underline{98.7} / \textbf{94.1} \\
% ExaGPT 
% & \underline{98.6} / \underline{92.3} 
% & \underline{98.9} / \underline{86.6} 
% & \textbf{99.5 / 96.0} 
% & \underline{99.6} / \underline{95.8} 
% & \textbf{99.2} / \underline{92.7} \\
% \bottomrule
% \end{tabular}
% }
% \caption{Performance Comparison with state-of-the-art non-interpretable detectors. The scores are AUROC / Acc@FPR=1\%. F.GPT: Fast-DetectGPT.}
% \label{sota}
% \end{table}

\paragraph{Interpretability of Detection Results.}
\label{interpretable_detectors}
To minimize the undesired consequences of detection, there is a need to develop a detector that provides interpretable evidence for its decisions. However, most detectors output only a binary label, and only a few studies aim to provide interpretable evidence.
GLTR \cite{gehrmann2019gltr} highlights tokens with high model likelihood. Other studies apply explainable ML techniques such as LIME and SHAP to supervised classifiers \cite{mitrovic2023chatgpt,wang-etal-2024-m4,lime16,shap17}.
DNA-GPT \cite{yang2023dnagptdivergentngramanalysis} compares $n$-gram overlaps between the target text and LLM-generated continuations, providing actual LLM texts with overlaps as evidence.

Unlike prior interpretable detectors, ExaGPT is grounded in the human decision-making process for verifying the origin of a text \cite{maurer06,barron-cedeno-etal-2013-plagiarism} and can provide more interpretable evidence, as explained in the previous sections.
% Specifically, ExaGPT compares a text with both human-written and LLM-generated texts from a datastore and investigates which class the text shares more similar spans with.
% As evidence of the detection, it provides similar span examples for each span in the text.
% The proximity of the span examples to each span in the text can help users judge how reliably correct the detection result is.

% When humans verify the origin of a text, they intuitively compare the text to other human-written and LLM-generated texts and investigate with which source it shares overlaps or rephrased similar spans \cite{maurer06,barron-cedeno-etal-2013-plagiarism}.
% Here, current detectors are not aligned with the human decision-making process and fail to present sufficiently interpretable evidence.
% Our work bridges this gap by proposing ExaGPT, an interpretable detection approach based on the human process of verifying the origin of a text.
% It compares the text with both human-written and LLM-generated texts from a datastore and investigates which class the text shares more similar spans with. Moreover, it can provide similar span examples for each span in the text as evidence. 
% The proximity of the retrieved spans to each span in the target text can help users judge how reliably correct the detection result is.

\paragraph{Example Retrieval for Interpretability.}
Beyond LLM text detection, presenting retrieved similar examples has contributed to improving the interpretability of models in various natural language processing tasks.
These tasks range from text generation, e.g.,~machine translation \citep{khandelwal2020nearest}, to sequential text classification, e.g.,~part-of-speech tagging \citep{wiseman-stratos-2019-label}, and grammatical error correction \citep{kaneko-etal-2022-interpretability}.
At each time step, these methods predict a token or a label from the output distribution of a base model interpolated with the distribution derived from retrieved nearest neighbor examples.

Our work has a similar direction of using retrieved similar examples for better interpretability with prior studies in other NLP tasks. In LLM text detection, it is particularly crucial to segment the target text into n-gram spans for better interpretability, with labels assigned individually \cite{cheng2025beyond}. Therefore, ExaGPT offers a unique mechanism that retrieves similar span examples for each span in the target text and optimizes the final span segmentation using dynamic programming.

\section{Conclusion}
% We introduced ExaGPT, an interpretable human vs. machine detection approach grounded in how humans verify the origin of a text. ExaGPT classifies a text by examining whether it shares more verbatim and semantically similar spans with human-written vs. with LLM-generated texts from a datastore. As evidence of the detection, ExaGPT provides similar span examples for each span in the text.
% Human evaluation and further analysis show that providing similar span examples allows users to judge detection correctness more effectively than prior interpretable detectors.
% Moreover, extensive experiments across multiple domains and generators demonstrate that ExaGPT achieves superior detection performance compared to previous strong detectors, even at a false positive rate of 1\%. 
% Overall, ExaGPT achieves both high interpretability and strong detection performance. 

We introduced ExaGPT, an interpretable human vs. machine detection approach grounded in the
human decision-making process of verifying the origin of a text. In particular, ExaGPT classifies a text by examining whether it shares more verbatim and semantically similar spans with human-written vs. with LLM-generated texts from an available datastore. As evidence of the detection, ExaGPT offers similar span examples for each span in the text. The human evaluation and further analysis show that providing similar span examples allows users to judge the correctness of the detection more effectively than prior interpretable detectors. Moreover, extensive experiments across various domains and generators revealed that ExaGPT has shown notably superior detection performance compared to previous strong detectors, even at a false positive rate of 1\%. These results indicate that ExaGPT is a detector with both high interpretability in its decision and high detection performance.

In future work, we plan to extend the human evaluation of interpretability to more diverse user populations and systematically explore more optimal datastore configurations to improve generalization across domains and generators.

\section{Limitations}
\paragraph{Scope of Human Evaluation.}
Our human evaluation relied on four annotators with NLP backgrounds, which may limit the generalizability of the results to typical users.
The evaluation was designed as a controlled comparison of interpretability between ExaGPT and existing detectors, rather than a general usability study. Since baseline methods require interpreting technical outputs (e.g., probability distributions or contribution scores), we used annotators with comparable expertise to ensure fair evaluation across methods.
Nevertheless, whether the interpretability benefits of ExaGPT extend to broader user populations remains unclear, and we leave this to future work.

\paragraph{Datastore Dependence.}
ExaGPT depends on the datastore for its detection by design. This raises concerns about its generalizability to text domains or generators not covered by the datastore.
Experiments show that the performance of ExaGPT is reduced in cross-domain and cross-generators settings. However, in practice, using multiple sources in the datastore, rather than relying on a single source, is considered more realistic. Our results confirm that, in such a setting, ExaGPT consistently maintained a reasonably high performance across multiple domains and generators. We look forward to future studies into optimal datastore configurations for more effective real-world deployment.

\section{Ethics and Broader Impact}
\paragraph{Human Subject Considerations.}
In our study, human subjects are engaged in identifying the correctness of the detection based on evidence. All annotators provided informed consent, were fully aware of the study's objectives, and had the right to withdraw at any time. 

\paragraph{Responsible Use of Detectors.}
It is extremely difficult to achieve perfect detection in the real-world. Given the severe consequences of misclassifications, detectors should provide evidence enabling users to assess the reliability of predictions and identify potential errors. ExaGPT improves interpretability by presenting similar spans as evidence. While such evidence can support human judgment, it does not eliminate the inherent uncertainty in detection. Therefore, more generally, detector outputs should not be treated as authoritative proof for sanctions, but rather as advisory signals considered alongside other contextual information.
% \paragraph{Transparency and Reproducibility.}
% To promote open research, we release our code and data to the public, including all human annotations.

\section{Acknowledgements}
This work was supported by JSPS KAKENHI Grant Number 25H01137. This work was supported by JST K Program Japan Grant Number JPMJKP24C3. This work was supported by JST SPRING, Japan Grant Number JPMJSP2106. These research results were obtained from the commissioned research (No.22501) by National Institute of Information and Communications Technology (NICT), Japan. We used ABCI 3.0 provided by AIST and AIST Solutions with support from ``ABCI 3.0 Development Acceleration Use''.

% Bibliography entries for the entire Anthology, followed by custom entries
%\bibliography{anthology,custom}
% Custom bibliography entries only
\bibliography{acl_latex}

\appendix
% \onecolumn
\section{Detailed Configurations of Baselines}
\label{configurations}
\paragraph{LR-GLTR.}
Following \citet{wang-etal-2024-m4}, we leverage the two categories of GLTR features: (1) the number of tokens in the top-\{10, 100, 1,000, 1,000+\} ranks in the predicted probability distribution of LLMs (four features), and (2) the probability distribution of the word divided by the maximum probability of any word at the same position over 10 bins between 0.0 and 1.0 (ten features).

\paragraph{DNA-GPT.}
For DNA-GPT, we set the truncation ratio $\gamma$ to 0.7 and 0.5, and the number of re-generations $K$ to 10 and 5 for closed-source and open-source LLMs. The \texttt{temperature} is the same as the one used to generate a target text, and the generation prompt is known. These configurations were found to ensure the favorable performance of DNA-GPT in \citet{yang2023dnagptdivergentngramanalysis}.  We set all other hyperparameters to their default values.

\section{Detection Evidence of Baselines}
\label{evidence_baselines}
\begin{figure*}[t]
 \begin{center}
  \centering\includegraphics[width=0.95\textwidth]{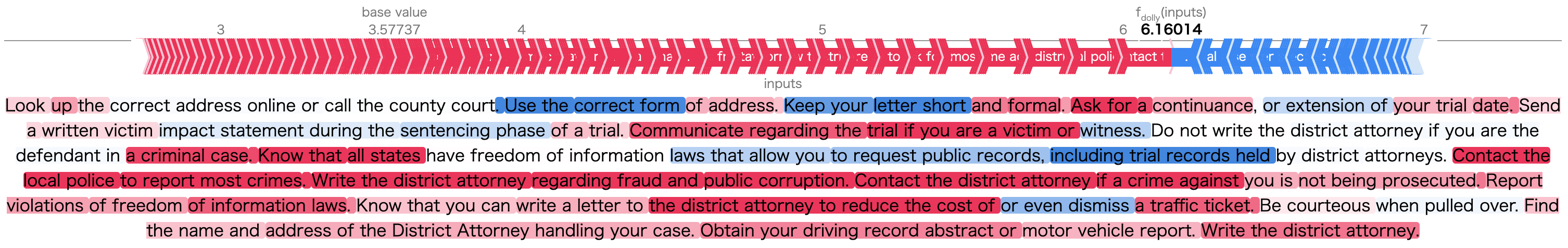}
  \caption{Example of evidence by RoBERTa with SHAP.}
  \label{shap}
 \end{center}
\end{figure*}

\paragraph{RoBERTa with SHAP.} 
Figure \ref{shap} depicts an example of evidence by RoBERTa with SHAP.
We visualize the evidence using the SHAP library\footnote{\url{https://shap.readthedocs.io/}}.
The red parts are spans that contribute to predicting LLM-generated. The blue parts are spans that contribute to predicting human-written. In the evidence, if the prediction value, $f({\rm inputs})$ moves further to the right compared to the base value (the expected value across all data samples), it is more likely to be LLM-generated.
When we hover over a colored part, we can see a score of how much the part contributes to the result. The more a span contributes to the decision, the darker its color.

\begin{figure*}[t]
 \begin{center}
  \centering\includegraphics[width=0.95\textwidth]{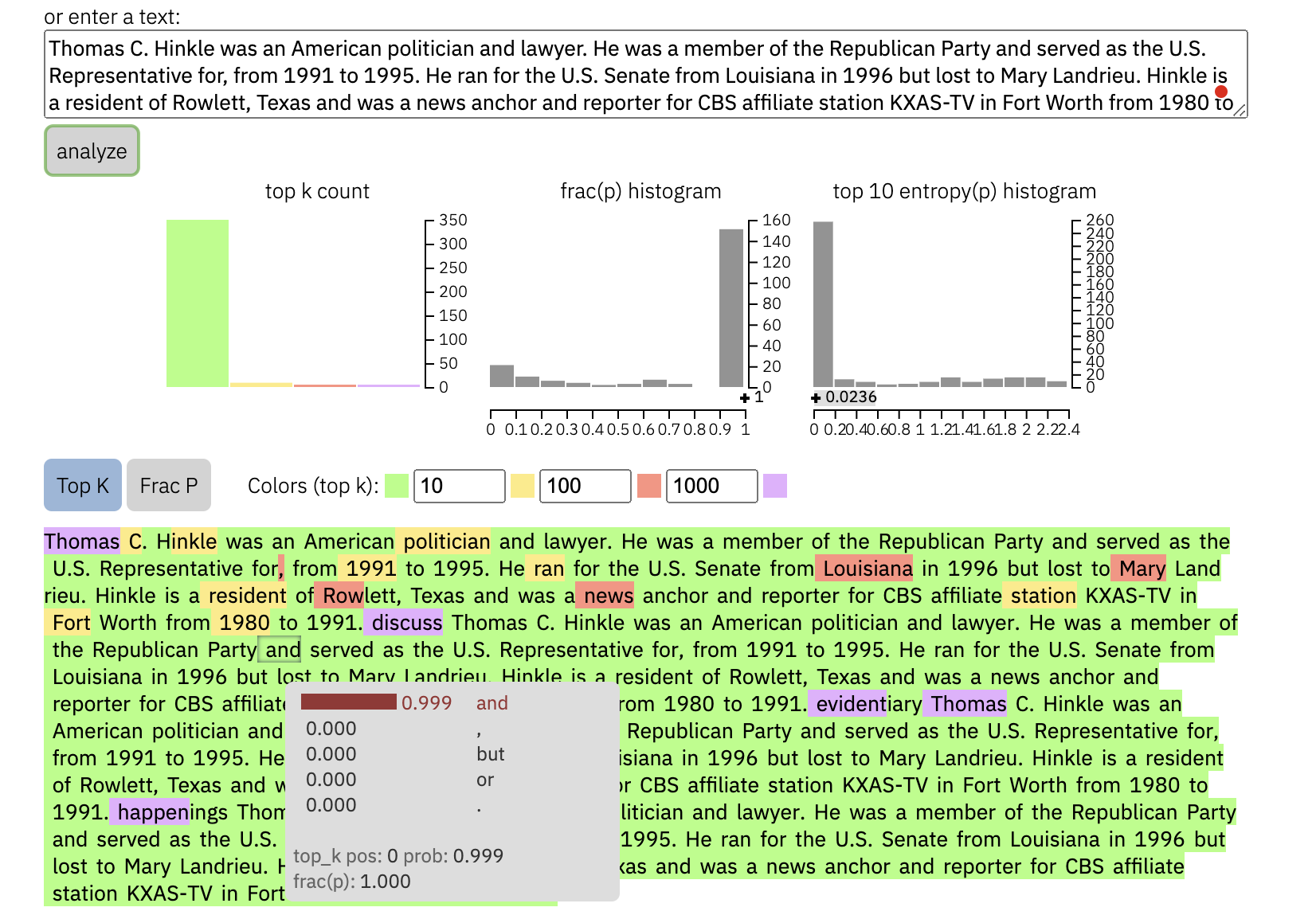}
  \caption{Example of evidence by LR-GLTR.}
  \label{gltr}
 \end{center}
\end{figure*}

\paragraph{LR-GLTR.}
Figure \ref{gltr} displays an example of evidence by LR-GLTR.
We leverage a demo app\footnote{\url{http://demo.gltr.io/client/index.html}} of GLTR, developed by \citet{gehrmann2019gltr}. 
It highlights tokens in different colors based on their rank of top-\{10, 100, 1,000, 1,000+\} in the predicted token distribution from an LLM. The higher the rank of the token, the more likely an LLM is to generate the token.
The green parts are spans that are most likely LLM-generated. The degree decreases in the order of green, yellow, red, and purple.
When we hover the cursor on a colored part, we can also see the predicted token distribution of an LLM.

\begin{figure*}[t]
 \begin{center}
  \centering\includegraphics[width=0.95\textwidth]{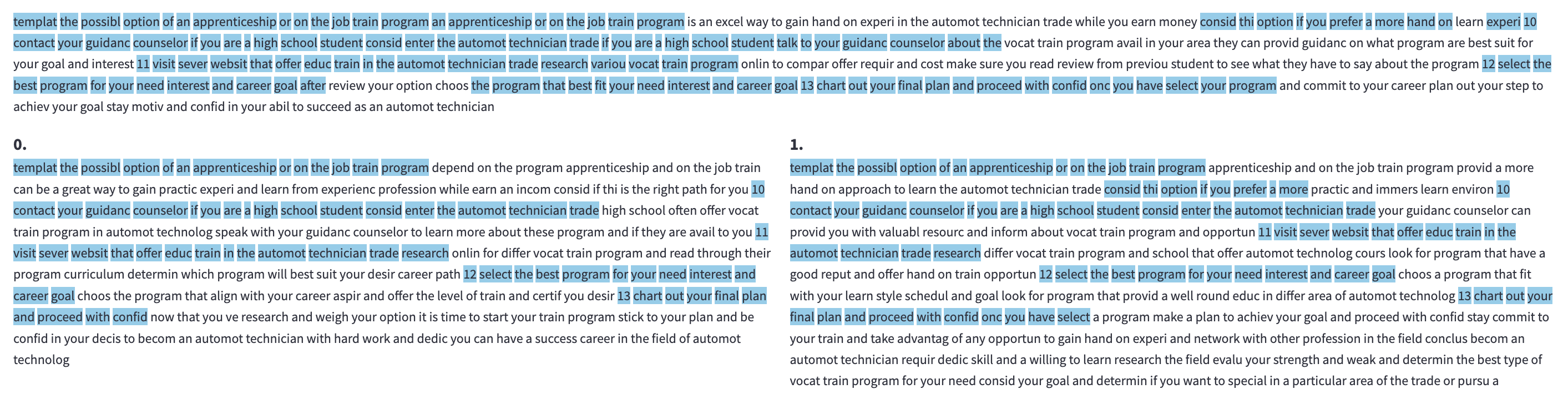}
  \caption{Example of evidence by DNA-GPT.}
  \label{dnagpt}
 \end{center}
\end{figure*}

\paragraph{DNA-GPT.}
Figure \ref{dnagpt} shows an example of evidence by DNA-GPT.
We implemented a demo app of DNA-GPT with the streamlit framework\footnote{\url{https://github.com/streamlit/streamlit}}.
It shows overlapped $n$-gram spans between a truncated target text and multiple LLM-generated continuations.
The more blue spans, the more likely the text is LLM-generated. For span matching, we follow the original implementation of DNA-GPT\footnote{\url{https://github.com/Xianjun-Yang/DNA-GPT}} where it was achieved by token-level matching based on preprocessing of the lower casing and stemming. We also set $n$ to 8 in order to show a large number of overlapped spans enough to interpret as evidence.

\section{Analysis Details}
\label{analysis_details}
% \paragraph{Example of \textit{k}-NN spans.}
% Table~\ref{knn_spans} presents examples of long spans (\textit{n} $=$ 19) with high reliability scores for a target span retrieved by ExaGPT.

\begin{figure*}[t]
 \begin{center}
  \centering\includegraphics[width=0.95\textwidth]{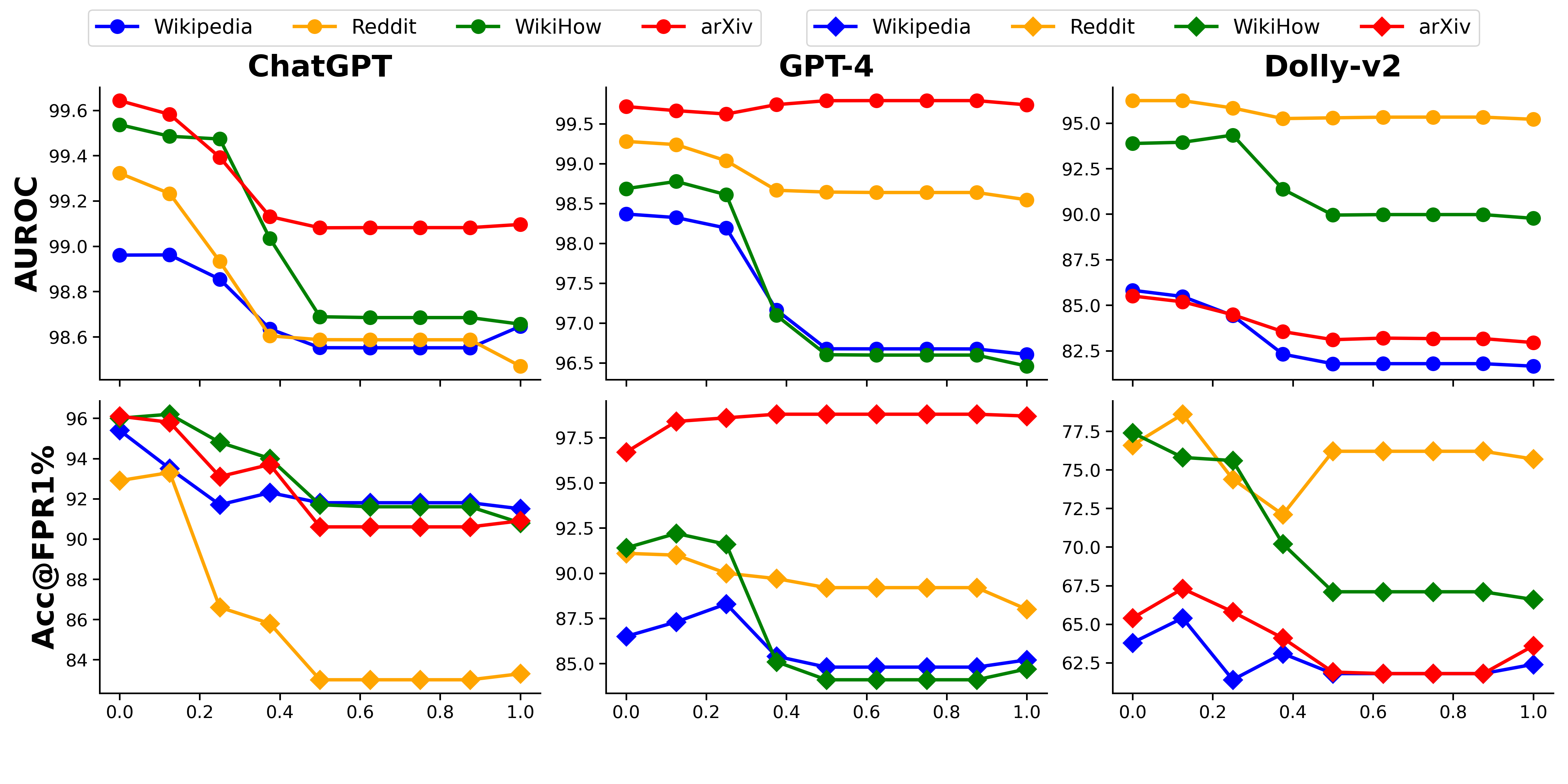}
  \caption{Impact of $\alpha$ on the detection performance of ExaGPT, including the AUROC and the accuracy at 1\% FPR, across four domains and three generators.}
  \label{alpha_effect_all}
 \end{center}
\end{figure*}

\paragraph{Impact of \texorpdfstring{$\alpha$}{Lg}.}
Figure \ref{alpha_effect_all} showcases the impact of $\alpha$ on the detection performance of ExaGPT across four domains and three generators.
We found similar overall trends of the impact of $\alpha$ in other LLMs, including GPT-4 and Dolly-v2, with the impact in ChatGPT, as explained in \S\ref{alpha}.

\begin{figure*}[t]
 \begin{center}
  \centering\includegraphics[width=0.95\textwidth]{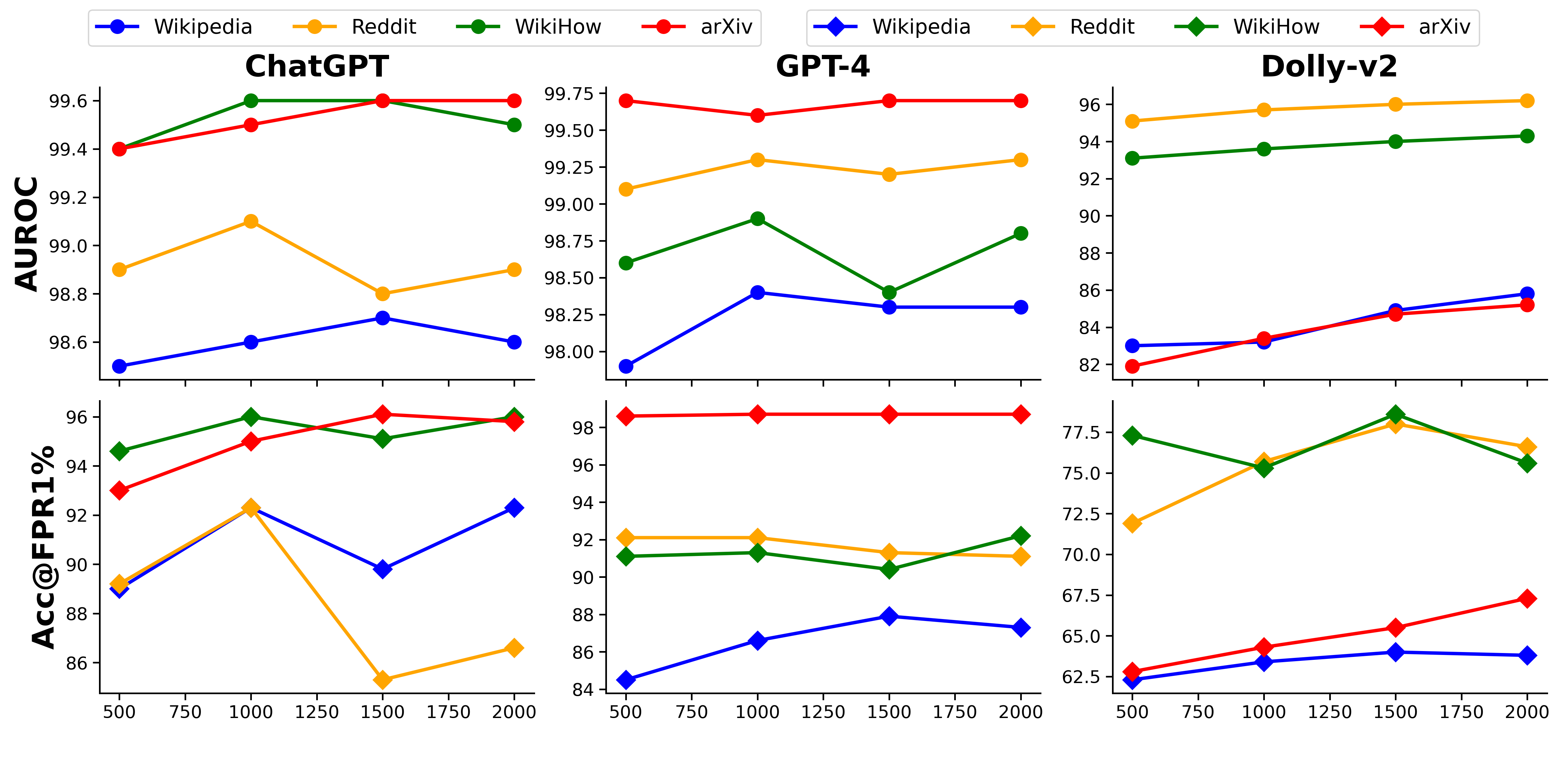}
  \caption{Impact of the datastore size on the detection performance of ExaGPT, including the AUROC and the accuracy at 1\% FPR, across four domains and three generators.}
  \label{datastore_size_effect_all}
 \end{center}
\end{figure*}

\paragraph{Impact of the Datastore Size.}
Figure \ref{datastore_size_effect_all} showcases the impact of the datastore size on the detection performance of ExaGPT across four domains and three generators.
We can observe similar overall trends of the impact of datastore size in other LLMs, including GPT-4 and Dolly-v2, with the impact in ChatGPT as explained in \S\ref{datastore}.

% \paragraph{Inference Cost.}
% In our preliminary analysis, we found that the inference cost of embedding generation and DP-based segmentation was negligible compared to the \textit{k}-NN search, which was the primary bottleneck due to the large datastore containing extensive \textit{n}-gram instances. Therefore, we have measured inference latency and GPU memory usage focusing on the \textit{k}-NN search component. We also have conducted experiments with FAISS using IVFPQ indexes, to reduce resource usage and improve inference speed.

% Table \ref{inference} provides the results on the WikiHow domain with ChatGPT. Here, \#Instance refers to the number of \textit{n}‑gram spans in the datastore. While we observe that achieving full performance with a datastore of 2,000 pairs requires considerable computational resources and higher latency, reducing the size to 500 pairs decreases GPU memory usage by 66\% and latency by 60\% without compromising detection performance. With the additional use of FAISS‑based \textit{k}-NN approximation, the requirements are further reduced by 87\% in memory and over 90\% in latency, while the performance drop is still moderate. These findings highlight the promising practical applicability of ExaGPT.

\section{Computational Budget}
We run all the experiments with two AMD EPYC 7453 CPUs and four NVIDIA A6000 GPUs. The total processing time is approximately 25 hours.

\end{document}